\documentclass[10pt, a4paper]{article}
\usepackage{lrec2022}
\usepackage{multibib}
\newcites{languageresource}{Language Resources}
\usepackage{graphicx}
\usepackage{tabularx}
\usepackage{soul}

\usepackage{epstopdf}
\usepackage[utf8]{inputenc}

\usepackage{hyperref}
\usepackage{xstring}

\usepackage{color}
\usepackage{tikz}
\usetikzlibrary{shapes,arrows}
\usepackage{subcaption}
\usepackage{tfrupee}  
\usepackage{algorithm2e}
\usepackage{amsmath}
\usepackage{adjustbox}
\usepackage{multirow}
\usepackage{multicol}
\usepackage{color, colortbl}

\usepackage{tabularx}
    \newcolumntype{L}{>{\raggedright\arraybackslash}X}
	
\definecolor{Gray}{gray}{0.9}

\title{What Does the Indian Parliament Discuss? \\An Exploratory Analysis of the Question Hour in the Lok Sabha
}

\name{Suman Adhya, Debarshi Kumar Sanyal}

\address{Indian Association for the Cultivation of Science \\
        Jadavpur, Kolkata -- 700032, India \\
        adhyasuman30@gmail.com, debarshisanyal@gmail.com\\}

\abstract{
The TCPD-IPD dataset is a collection of questions and answers discussed in the Lower House of the Parliament of India during the Question Hour between 1999 and 2019. Although it is difficult to analyze such a huge collection manually, modern text analysis tools can provide a powerful means to navigate it.  In this paper, we perform an exploratory analysis of the dataset. In particular, we present insightful corpus-level  statistics  and a detailed analysis of three subsets of the dataset. In the latter analysis, the focus is on understanding the temporal evolution of topics using a dynamic topic model.  We  observe that the parliamentary conversation indeed mirrors the political and socio-economic tensions of each period. 
 \\ \newline  }

\begin{document}

\maketitleabstract

\section{Introduction}
The Parliament of India is the highest legislative body of India. The members of its Lower House or the Lok Sabha are directly elected by the people while its Upper House comprises representatives elected by the members of all State Legislative Assemblies. 
Although parliamentary proceedings are immensely useful to a political scientist, they are too large to be manually analyzed. This motivates the use of algorithmic tools to explore them.
In this paper, we analyze the TCPD-IPD dataset \citelanguageresource{tpcd} of around 298K pairs of questions and answers (QA) in English discussed in the Lok Sabha during the Question Hour -- the first hour of every business day of the Parliament -- from 1999 to 2019 spanning four Lok Sabha terms (13th term: 1999-2004, 14th: 2004-09, 15th: 2009-14, 16th: 2014-19). 
During the Question Hour, any Member of Parliament in the Lok Sabha (abbreviated: MP) may ask any question to the ministers related to the administrative activity of the government, and thus,  hold it accountable for its actions \cite{sanyal2016regulating,tripathi21pandemic}. Question time is also an integral part of many other parliamentary democracies like those of Canada, Australia, and the UK \cite{martin2014roles}. 

Technical specification of the TCPD-IPD dataset appears in \cite{saloni2019tpcd}. But the dataset has not been explored, except in \cite{sen2019studying} where the authors aim to identify, for a few chosen themes, whether the questions asked by MPs echo the trend in mass media and social media. Topic modeling has been used to analyze the parliamentary proceedings of various countries, see, e.g., \cite{greene2017exploring,gkoumas2018exploring,ishima2020electoral}. 
In this paper, we study TCPD-IPD using the following pipeline. First, a static topic model of the entire dataset is built and the top topics are identified. Then a subset of the dataset is selected for further analysis as follows: (a) A dynamic topic model is built on it; (b) The temporal evolution of topics and words in a topic are plotted; (c) The top-ranking documents at a given time in each of these plots are analyzed. We obtain interesting insights from this analysis. This demonstrates the effectiveness of our technique. Our specific contributions are: 
\begin{enumerate}
    \item We describe important high-level statistical features of the dataset and bias in the participation of MPs (Section  \ref{sec:dataset}).
    \item We identify the top topics in the entire dataset (Section \ref{sec:globalTopics}).
    \item We make a more nuanced study of the QA pertaining to three specific ministries -- Finance, Railways, and Health and Family Welfare --- by building a dynamic topic model in each case (Section \ref{sec:ministrywise}).
    \item For each ministry mentioned above, we identify  words that showed significant variation in their probability in a topic over time (Section \ref{sec:ministrywise}) and the major events to which they relate. Thus, word choreography in a dynamic topic model is used to reconstruct events in political history. We hope the insights and lessons from the past will help inform future responses to critical national issues.
\end{enumerate}

\section{High-Level Features of TCPD-IPD}
\label{sec:dataset}
We enumerate below some interesting insights we obtained from the dataset.
    \subsection{Ministry-wise data distribution}
\begin{figure} [!htbp]
        \centering
        \includegraphics[width=1\linewidth]{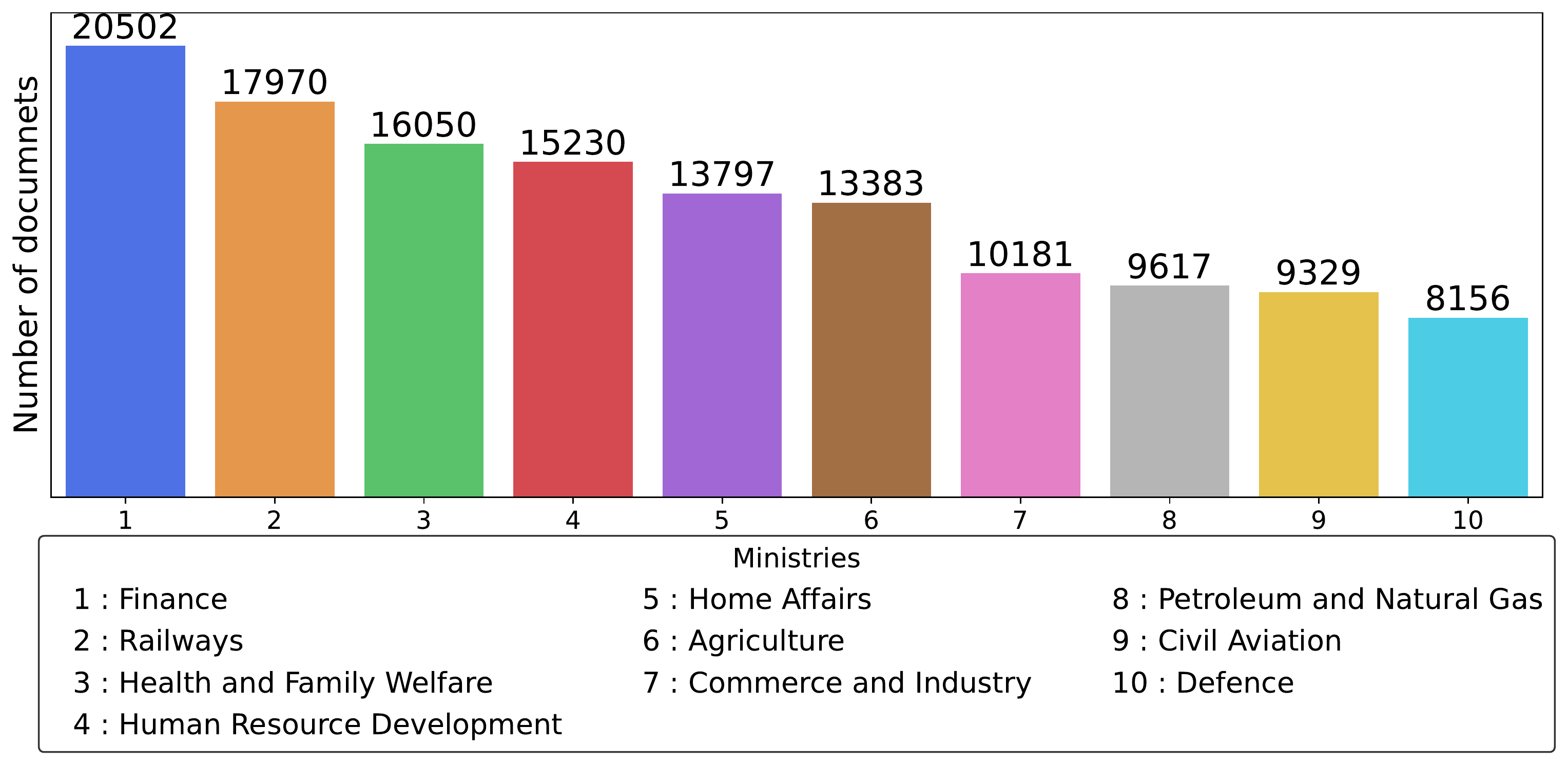}
        \caption{Ministry-wise data distribution.} 
        \label{fig:EDA_dataDist_t10}
        \end{figure}

    TCPD-IPD contains questions related to 85 different ministries. Figure \ref{fig:EDA_dataDist_t10} shows the data distribution for the top ten ministries (comprising almost $50\%$ of the full dataset). Clearly, \textit{Finance}, \textit{Railways} and \textit{Health and Family Welfare} are the top three ministries.
    \subsection{Term-wise data distribution}
    Among a total of 298,292 questions, the number of questions asked in each term of the Lok Sabha is as follows: 13th: 73,531; 14th: 66,371; 15th: 79,401; 16th: 78,989. We found out that over the span covered in this dataset, Lok Sabha always met for more than 50 sittings in a year, except in 2004 (48 sittings) and 2008 (46 sittings). This correlates with the lowest number of questions asked in 2004 (9398 questions) and 2008 (9851 questions). Two of the three sessions in 2004 and all sessions in 2008 are included in the 14th Lok Sabha. Although the fewer sittings are highlighted in many news reports \cite{varma2020low}, we did not find mention of its impact on Question Hour, which we clearly identified  above.
    \subsection{Participation of MPs}
    Unlike the previous Lok Sabha terms, in the 16th  the ruling alliance asked more questions than the opposition. In that term, they had a historic $65\%$ share of the House.
    Since the numeric strength of the ruling alliance is, by rule, higher than that of the opposition, we normalize them to get an idea of participation from each side \textit{had the number of representatives from either side been equal}. 
    We believe this will afford a fairer comparison between their participation. Let $R_n$ and $O_n$ be the number of members of the ruling alliance and the opposition, respectively, and $R_q$ and $O_q$ be the number of questions asked by the ruling alliance and the opposition, respectively. Then $R_{pp} = \left( R_q \times \frac{O_n}{R_n + O_n} \right)$, $O_{pp} = \left( O_q \times \frac{R_n}{R_n + O_n}\right)$. 
    As seen in Figure 2, in the transformed space, the opposition still asks more questions than the ruling alliance,  matching the expectations of a healthy democracy. 
    
    \begin{figure}
    \centering
    \includegraphics[width=1\linewidth]{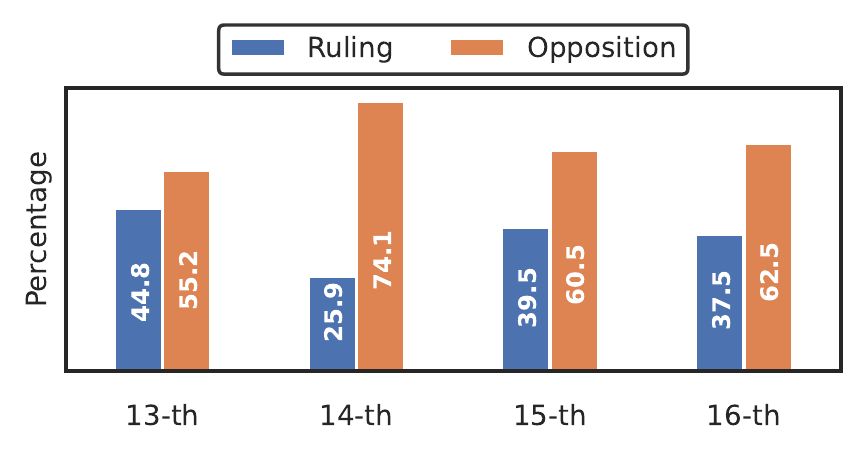}
    \captionsetup{justification=centering}
    \caption{Participation of ruling alliance vis-a-vis that of the opposition.} 
    \label{fig:EDA_RvO}
    \end{figure} 
    
    \subsection{Gender and caste bias in participation} 
\begin{figure}[!htbp]
    \centering 
    \begin{subfigure}[b]{0.49\linewidth}
        \includegraphics[width=\textwidth]{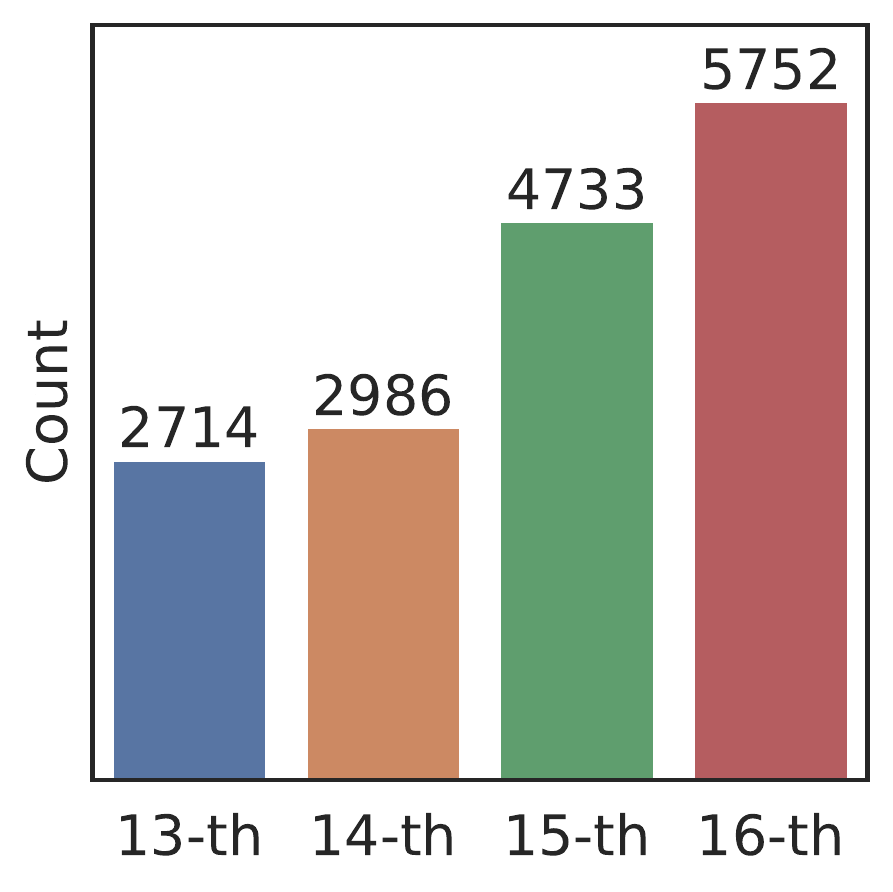}
        \caption{\textit{Gender}-related.}
        \label{fig:Gender_bias}
    \end{subfigure} \hfill
    \begin{subfigure}[b]{0.49\linewidth}
        \includegraphics[width=1\textwidth]{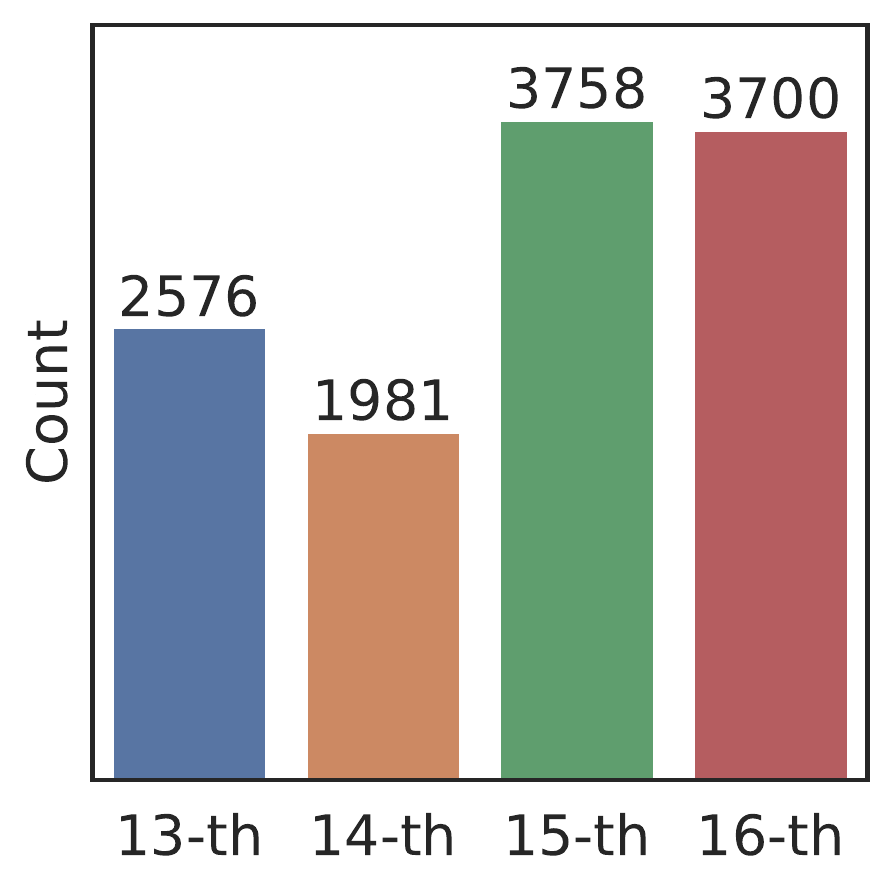}
        \caption{\textit{Caste}-related.}
        \label{fig:Caste_bias}
    \end{subfigure}
    \caption{\textit{Gender} and \textit{caste} related discussions in each Lok Sabha term.}
    \label{fig:Bias_analysis}
\end{figure}

    Over the four Lok Sabha terms covered by the dataset, 91.6\% of questions were asked by men while 8.4\% of questions were raised by women. The average gender ratio of men to women was $8.3 \!\!:\!\! 1$ over the same four terms. Thus the skewed gender ratio correlates with the distribution of questions. It is noteworthy here that the Women's Reservation Bill proposing the reservation of one-third of the seats in the Lok Sabha for women has been pending since 2010 \cite{marwah2019gender}, thus, allowing the bias to continue. As regards the caste distribution, 80.6\% of questions were raised by MPs from the general caste while the rest come from the reserved categories. Note that 24.03\% of Lok Sabha seats are reserved for the reserved categories while the rest belong to the general caste. Figure \ref{fig:Bias_analysis} shows the number of questions on gender and caste-related issues asked in the Parliament;  Appendix \ref{ap:bias} lists the keywords we used for this  analysis.
    
\section{Topic Model for TCPD-IPD}
\label{sec:globalTopics}
\begin{figure}
\centering
\includegraphics[width=0.96\linewidth]{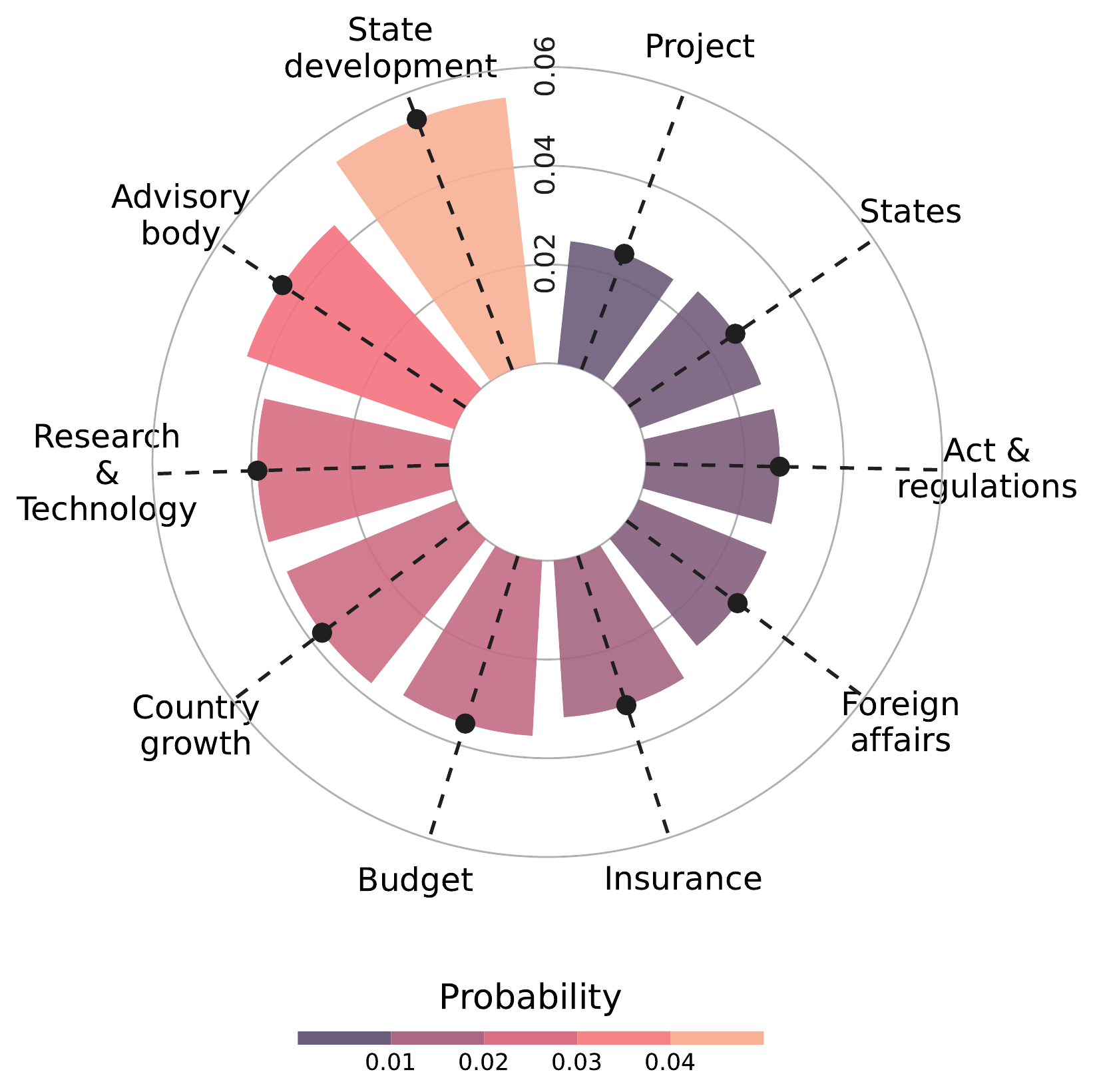}
\captionsetup{justification=centering}
\caption{Hot topics discussed in the Indian parliamentary QA sessions.} 
\label{fig:HotTopics}
\end{figure}

We used topic modeling to get a thematic view of the TCPD-IPD dataset. Researchers have observed that Latent Dirichlet Allocation (LDA) (\cite{blei2003latent}), which employs Gibbs sampling for inference, often produces superior topics than those from modern variational inference-based topic models; see, e.g.,  \cite{blei2017variational,lisena2020tomodapi}. This  motivated us to use LDA instead of neural topic models. We pre-processed the entire TCPD-IPD corpus, used LDA to extract 50 topics, and manually labeled them (See Appendix \ref{ap:preprocess}-\ref{ap:lda}).
We filtered out a few noisy, heterogeneous topics and among the rest plotted the top ten topics in Figure  \ref{fig:HotTopics}. Clearly, there is a huge emphasis on growth in the economy and science, at the state and national levels. 
\section{Ministry-wise Analysis}
\label{sec:ministrywise}
We have selected three ministries -- Finance, Railways, and Health and Family Welfare -- for further analysis. We set the topic count to 20 for all three ministries. 
To model the temporal variation in topics, we had to make a choice between the classical Dynamic Topic Model (\cite{blei2006dynamic}) which is essentially an adaptation of LDA  for sequential data (hereafter, called LDAseq) and the more modern Dynamic Embedded Topic Model (D-ETM) (\cite{DETM}), both developed by David Blei and collaborators.  
In our experiments, we found LDAseq produced better topics (see Appendix  \ref{ap:quality}). 
A similar observation is reported made in \cite{DETM}.  

\begin{figure*} [!htbp]
        \centering
        \includegraphics[width=0.99\linewidth]{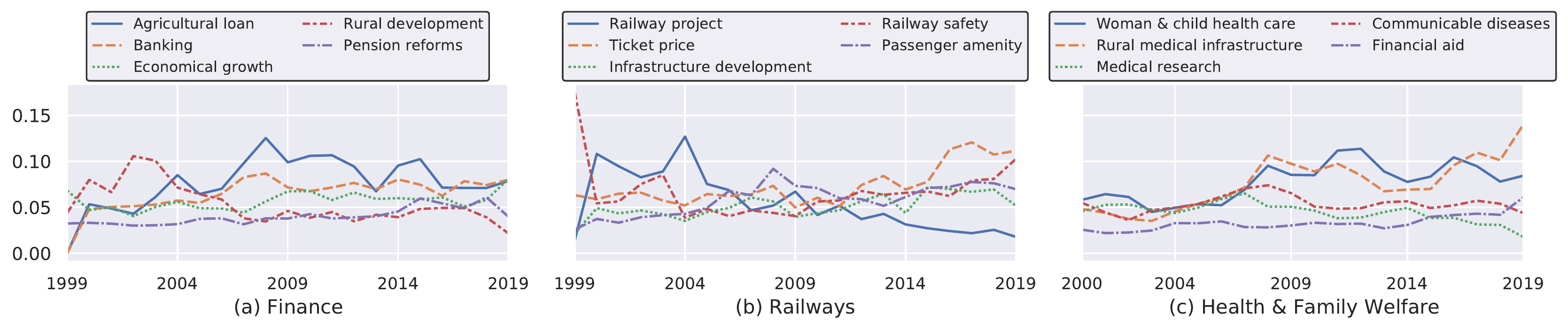}
        \caption{Temporal evolution of topics in 3 subsets of TCPD-IPD using LDAseq.} 
        \label{fig:LDAseq_Finance_full_line}
\end{figure*}

Using LDAseq, we obtained the temporal evolution of the probabilities of five topics for each ministry, as shown in 
Figure \ref{fig:LDAseq_Finance_full_line}. In Finance, the peaks in `agricultural loan' in 2008 and 2014 relate to the debt waivers announced at that time, though farmers' loans remain a perennially important topic. The focus on `rural development' slowly reduces while other topics like `banking', `economic growth' (more questions asked when GDP change is unexpectedly high or low), and `pension schemes' remain more stable.
In the Railways, the early 2000s witnessed many new government projects and associated parliamentary QA, but the focus gradually shifted to `infrastructure development' and `passenger amenity', where questions veered around the increasing private participation. The announcement of many projects on rail safety in the last Lok Sabha term is indicated by an increased presence of the topic `railway safety'.
The steep price hike in passenger and freight fares in 2014 sparked intense deliberation in the Parliament.  
In the Ministry of Health and Family Welfare, discussions on women and child care and rural medical infrastructure increased after the National Health Mission was launched in 2005. The focus on medical research in the early part of the decade led to the establishment of many premier medical institutions throughout the country but gradually the interest waned.

The \textit{rare} words in a topic are often more informative and captured specific events or issues. So in the following sub-sections, we choose a few representative topics from each ministry and show the temporal evolution of the probability of the rare words in the selected topics. In each plot, we also annotate one of the dominant words with example questions asked during the Question Hour.

\subsection{Finance}
\label{sec:finance}
\begin{figure*}[!htbp]
    \centering 
    \begin{subfigure}[b]{0.49\textwidth}
        \includegraphics[width=\textwidth]{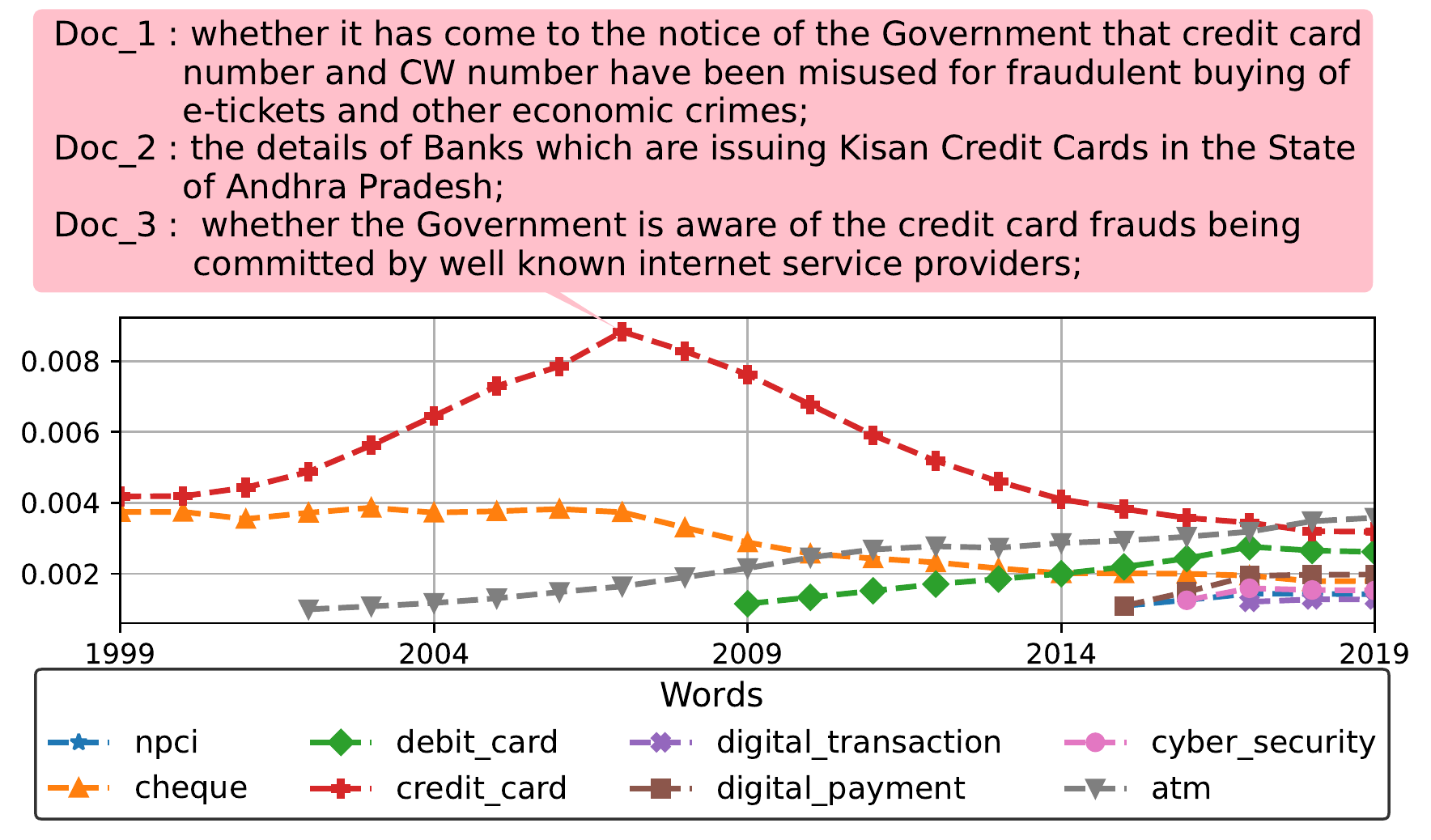}
        \caption{Topic: \textit{Banking}}
        \label{fig:LDAseq_Finance_full_T2}
    \end{subfigure} \hfill
    \begin{subfigure}[b]{0.49\textwidth}
        \includegraphics[width=1\textwidth]{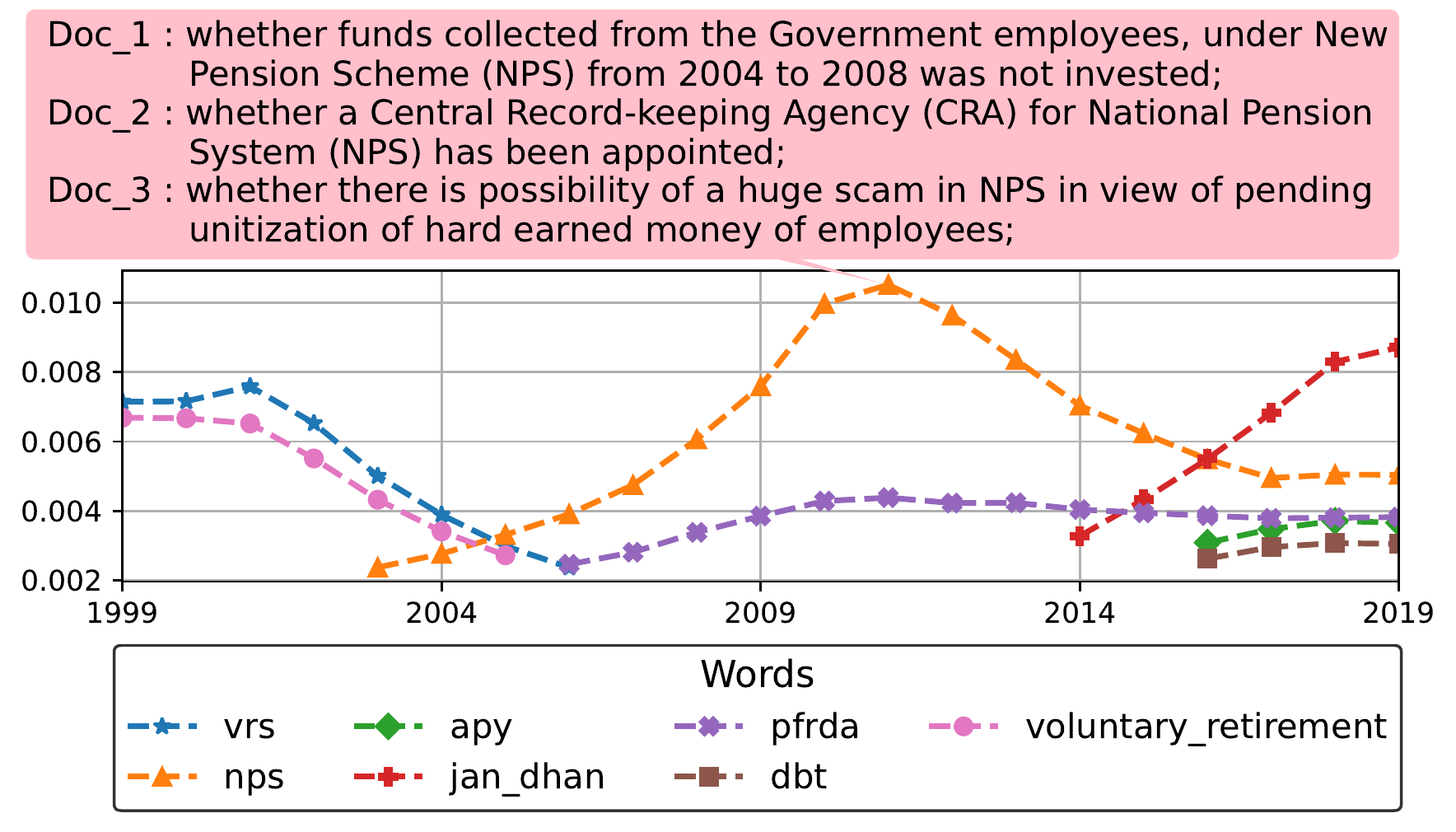}
        \caption{Topic: \textit{Pension Reforms}}
        \label{fig:LDAseq_Finance_full_T8}
    \end{subfigure}
    \caption{Selected topics obtained by running LDAseq on the Finance subset of TCPD-IPD.}
    \label{fig:LDAseq_Finance_full_side}
\end{figure*}
We have selected two topics, \textit{`banking'} and \textit{`pension reforms'}, and plotted the probability of a few selected tokens in them as a function of time in Figure  \ref{fig:LDAseq_Finance_full_side}. 
In the topic `banking' shown in Figure \ref{fig:LDAseq_Finance_full_T2}, we find that the word `credit\_card' peaks in 2007. Our analysis shows that most of the questions around this time are related to the growing credit card frauds in India and the sudden rise in credit card interest rates by some banks, coinciding with reports in mass media. Another aspect of credit card-related discussion is \textit{Kisan Credit Card} (KCC) -- a low-interest credit card for farmers, which was introduced in 1998 and significantly improved in 2004.
The steady rise in discussion on debit cards correlates with the increasing adoption of debit cards (that avoided the debt trap of credit cards) in India. With demonetization and increased government emphasis on end-to-end digital -- as opposed to cash --  transactions, terms like `atm', `digital\_transaction', and `cyber\_security' gain prominence while the popularity of more traditional mediums like `cheque' reduces.

Figure \ref{fig:LDAseq_Finance_full_T8} shows the topic of pension reforms.  
In late 2003, the Government of India notified that it was abolishing the then-existing government-funded pension system for all its new employees 
and that they would come under the National Pension System (NPS) to be administered through the new  Interim Pension Fund Regulatory and  Development Authority (PFRDA). NPS enabled subscribers to make planned savings for post-retirement income. NPS was extended to all Indian citizens in 2009.  
Being a monumental change, NPS provoked a number of questions that peaked around 2011; MPs wanted to know the details of the scheme, including its performance, implementation challenges, extension to unorganized sectors, and even its security. The government introduced the Voluntary Retirement Scheme (VRS) in nationalized banks in the early 2000s to reduce the financial load on the public exchequer. Thousands of employees across various  organizations accepted VRS in 2000-2001. Given the high unemployment rate in the country, VRS generated a lot of panic among  people and pointed questions in the Parliament on the government's future plans for its workforce. 
The other visible terms `apy', `dbt', and `jan\_dhan' refer to  recent financial inclusion programs.
       
\subsection{Railways}
\label{sec:railways}
\begin{figure}[!htbp]
  \centering 

        \includegraphics[width=0.99\linewidth]{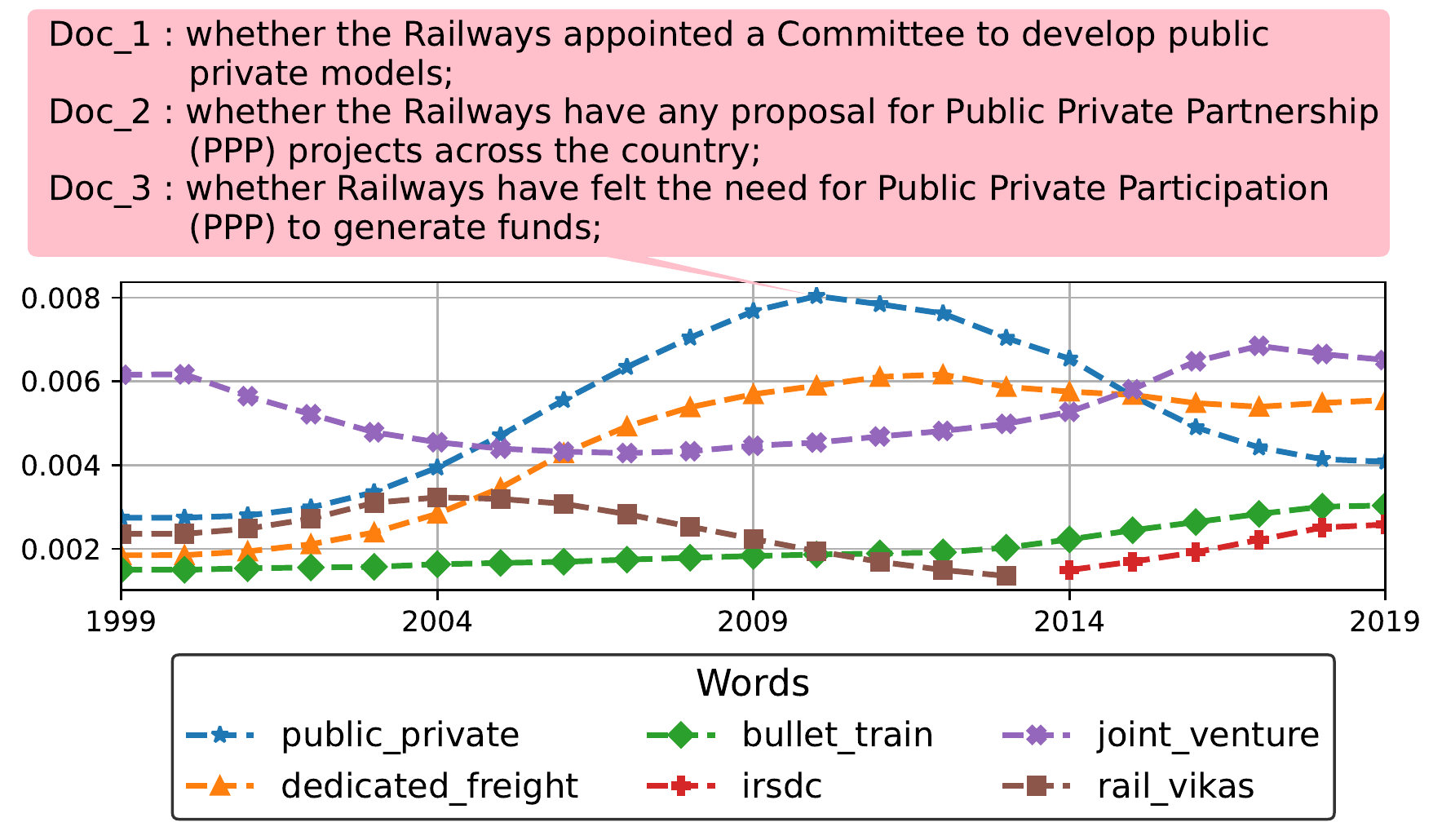}
        \caption{Topic \textit{Infrastructure Development} in Railways.}
        \label{fig:LDAseq_Railways_full_T12}
\end{figure}

Here, we highlight only one topic, namely, `infrastructure development' which is displayed in Figure  \ref{fig:LDAseq_Railways_full_T12}. 
Observe the peak of the term `public\_private' around 2010 when the Railways ministry introduced a new model for a public-private partnership to modernize the Indian Railways.  Indeed, there have always been questions and panic in the Lok Sabha on the public-private models as  privatization of the economy could increase fares, job loss, and casualization of labor  \cite{makhija2006privatisation,reddy2019privatisation}. A related term `joint\_venture' was a part of many discussions. While early uses of it (in the 2000s) focused on joint ventures of Indian Railways with other public sector companies, the recent focus (since 2015) has been on the increasing role of private players. 
Indeed similar exchanges occurred between the MPs and the Civil Aviation Ministry on the privatization of airlines. 
Terms like `rail\_vikas' and `irsdc' refer to companies owned by Indian Railways and entrusted with the maintenance of Railways. The rise in freight volumes led to the ideation of Dedicated Freight Corridors (DFC) in 2005 
and generated a number of questions (`dedicated\_freight') related to their cost,  progress, and expansion. 
Discussions on the introduction of bullet trains have been present for a long time but they gathered momentum when a vision document was tabled by the government in December, 2009 and the construction of the Mumbai-Ahmedabad high-speed rail corridor started in 2017.

\subsection{Health and Family Welfare} 
\label{sec:health}
\begin{figure}[!htbp]
    \centering 
    \includegraphics[width=0.99\linewidth]{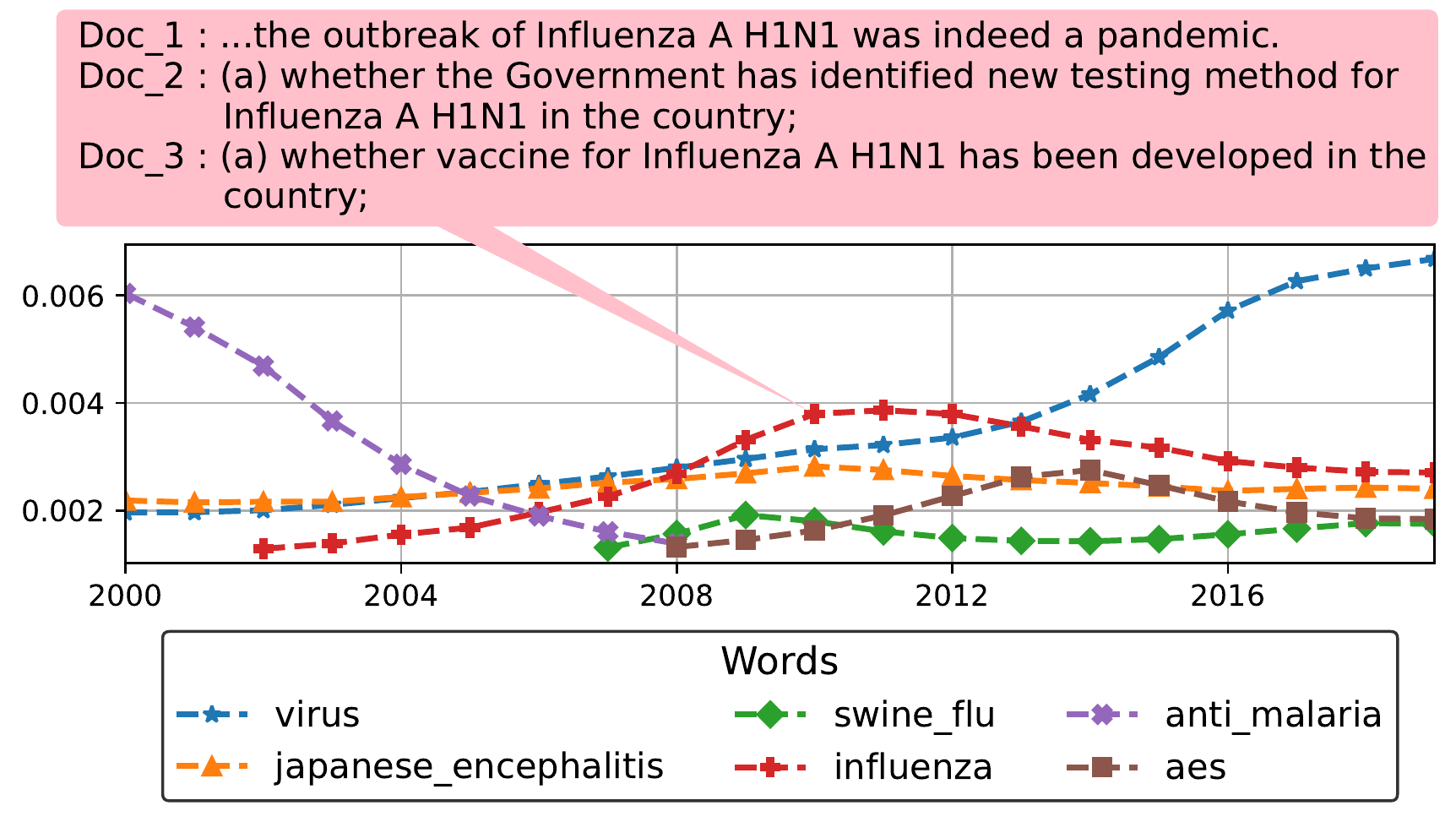}
    \captionsetup{justification=centering}
        \caption{Topic \textit{Communicable Diseases} in Health and Family Welfare.}
        \label{fig:LDAseq_Health_full_T13}
\end{figure}

Here, we have chosen the topic `Communicable Diseases'. Figure \ref{fig:LDAseq_Health_full_T13} shows the evolution of word probabilities in this topic. Clearly, we find an increasing focus on the word `virus' because India has been repeatedly hit by the Swine influenza virus (such as H1N1), including in 2009 when swine-flu turned into a pandemic and in 2014-15 \cite{kshatriya2018lessons}. MPs in the Lok Sabha enquired about the number of cases, test administration, government interventions, vaccination drives, the role of WHO, and the effort to develop indigenous vaccines. 
Another major disease in India has been malaria but a response in Lok Sabha informs us that malarial death reduced steadily over the years, and that is attested to by the steady decline in its focus in the Parliament. India recorded thousands of deaths due to Acute Encephalitis Syndrome (AES) in 2008-14 \cite{ghosh2016acute} and Japanese Encephalitis in 2005-11 \cite{adhya2013japanese}. During these unfortunate occurrences,  the Lok Sabha witnessed intense discussions on diseases.

\section{Conclusion and Future Work}
We identified the salient features of TCPD-IPD and then illustrated the temporal evolution of topics in three important subsets of the data. 
In the future, we will attempt to automatically detect topical change points, annotate them with trigger events (e.g., VRS announcement), auto-summarize the top documents containing a specific topic or word at a given time,  and motivate investigative reporting or research on the impact of the most sensitive topics discussed in the Parliament (e.g., the effect of VRS on mid-age employees). We hope our study will help  construct more probing parliamentary questions and formulate better national policies.

\section{Appendix}
\label{sec:appendix}
 \subsection{Data preprocessing for topic modeling}
 \label{ap:preprocess}
 We have removed the punctuation from the dataset, then lowercased and lemmatized the words. We have also removed the stopwords and filtered out the remaining words having document frequency lower than $0.001$ and higher than $0.95$. We have only kept the words that have at least 3 characters. Finally, we removed the documents with less than 3 words in them. After preprocessing, we created bigrams to better capture word co-occurrence statistics.
 
 \subsection{Configuration for topic models}
 \label{ap:tm}
 We used the following hyperparameters to run LDA, LDAseq, and D-ETM. 
  \begin{enumerate}
      \item \textbf{LDA} \cite{blei2003latent}: We use the Gensim implementation of the \href{https://radimrehurek.com/gensim/models/ldamodel.html}{LDA model}. To enable reproducibility, we use a fixed random seed, i.e., set the \texttt{random\_state} $=2021$. We set \texttt{passes} to $20$ and use the default values for the remaining hyperparameters.
      \item \textbf{LDAseq} \cite{blei2006dynamic}: We use the \href{https://radimrehurek.com/gensim/models/ldaseqmodel.html}{Gensim implementation}. We set the \texttt{random\_state} value as $2021$ and \texttt{passes} as $20$. For the rest of the  hyperparameters, we use the defaults.
      \item \textbf{D-ETM} \cite{DETM}: We use the original  implementation\footnote{ \url{https://github.com/adjidieng/DETM}} with \texttt{batch\_size} $=64$ and \texttt{epochs} $=100$, and keep the default values for the rest of the hyperparameters.
  \end{enumerate}

\subsection{Topics in TCPD-IPD}
 \label{ap:lda}
\begin{figure} [htbp]
        \centering
        \includegraphics[width=0.95\linewidth]{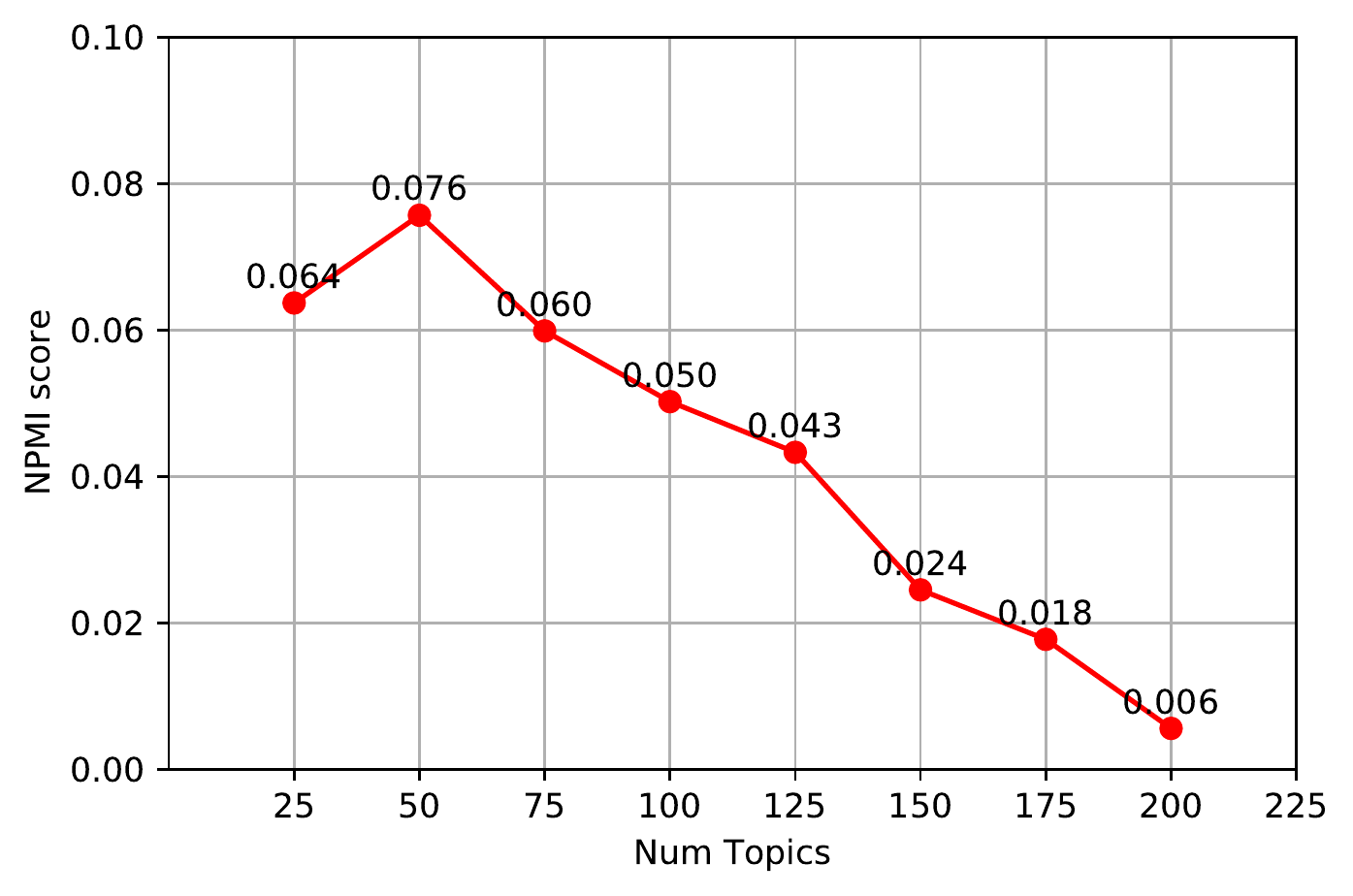}
        \captionsetup{justification=centering}
        \caption{NPMI values for the different number of topics in LDA over the full TCPD-IPD dataset.} 
        \label{fig:LDA_Global_NPMI}
\end{figure}

 We used LDA to extract $50$ topics from the entire TCPD-IPD dataset as it achieved the highest coherence score (see Figure \ref{fig:LDA_Global_NPMI}) when topic count was varied from $25$ to $200$ in steps of $25$.
The topics with their manual labels are shown in Table \ref{Tab:Global_topics}. Note that we could not run LDAseq because it was taking too much time. It did not terminate even after running for 72 hours. We could not run D-ETM as it experienced the posterior collapse, which could not be resolved even after changing the hyperparameters. 
\begin{table}[!htbp]
\centering
\begin{adjustbox}{width=0.99\linewidth}
  \begin{tabular}{|c|c|} \hline 
    \textbf{Top-5 words in topic} & \textbf{Manual label} \\ \hline
        country, foreign, agreement, international, sign & \textit{Foreign affairs} \\ \hline
        standard, quality, safety, use, pollution & \textit{Pollution} \\ \hline
        urban, city, housing, delhi, construction & \textit{Urban development} \\ \hline
        security, state, police, home, affair & \textit{Law \& order} \\ \hline
        road, highway, national, construction, state & \textit{Road construction} \\ \hline
        bank, loan, rbi, credit, account & \textit{Banking} \\ \hline
        committee, state, review, recommendation, report & \textit{Advisory body} \\ \hline
        state, andhra\_pradesh, uttar\_pradesh, maharashtra, tamil\_nadu & \textit{States} \\ \hline
        fund, crore, release, year, state & \textit{Budget} \\ \hline
        gas, oil, milk, production, natural\_gas & \textit{Petroleum} \\ \hline
        air, airport, defence, civil\_aviation, airline & \textit{Civil aviation} \\ \hline
        post, employee, central, office, officer & \textit{Central Employee} \\ \hline
        export, import, product, trade, textile & \textit{Market overview} \\ \hline
        project, complete, work, cost, sanction & \textit{Project} \\ \hline
        case, court, high\_court, person, disability & \textit{Judicial system} \\ \hline
        solar, renewable\_energy, energy, system, power & \textit{Renewable energy} \\ \hline
        act, state, provision, section, rule & \textit{Act \& regulations} \\ \hline
        service, information, provide, telecom, network & \textit{Information technology} \\ \hline
        payment, tax, pay, revenue, amount & \textit{Tax} \\ \hline
        power, plant, capacity, supply, state & \textit{Electricity plant} \\ \hline
        sport, fertilizer, ltd, limit, corporation & \textit{Fertilizer plant} \\ \hline
        madam, action, complaint, case, report & \textit{Grievance} \\ \hline
        water, river, state, resource, drinking\_water & \textit{Water resources} \\ \hline
        education, school, university, student, human\_resource &  \textit{HR in Education} \\ \hline
        scheme, state, development, provide, implement & \textit{State development} \\ \hline
        coal, mine, production, mineral, mining & \textit{Coal mining} \\ \hline
        rural, district, area, village, functional & \textit{Rural issues} \\ \hline
        delhi, mumbai, city, gujarat, chennai & \textit{Metropolitan areas}  \\ \hline
        health, state, family\_welfare, drug, medical & \textit{Health services} \\ \hline
        answer, lok\_sabha, reply, statement, lay & \textit{Parliament} \\ \hline
        farmer, agriculture, crop, agricultural, production & \textit{Agriculture} \\ \hline
        china, bangladesh, island, nepal, disaster & \textit{Natural disaster} \\ \hline
        increase, year, rate, country, reduce & \textit{Country growth} \\ \hline
        steel, port, connectivity, capacity, major & \textit{Ports shipping} \\ \hline
        tribal, schedule, minority, scholarship, tribe & \textit{Tribal affairs} \\ \hline
        food, price, consumer, state, foodgrain & \textit{Food price} \\ \hline
        sector, private, policy, public, investment & \textit{Private sector} \\ \hline
        woman, child, employment, worker, labour & \textit{Woman \& child labour} \\ \hline
        ngo, bihar, society, organisation, organization & \textit{NGO} \\ \hline
        tourism, culture, site, tourist, development & \textit{Tourism} \\ \hline
        industry, development, infrastructure, unit, scheme & \textit{Industry} \\ \hline
        company, issue, guideline, application, insurance & \textit{Insurance} \\ \hline
        research, technology, centre, training, national & \textit{Research \& Technology} \\ \hline
        railway, train, station, passenger, rail & \textit{Railway} \\ \hline
        land, forest, area, environment\_forest, state & \textit{Wildlife conservation} \\ \hline
        state, proposal, set, chhattisgarh, propose & Miscellaneous \\ \hline
                vehicle, procurement, website, award, contract & Miscellaneous \\ \hline
        year, wise, state, last\_three, number & Miscellaneous \\ \hline
        due, pleased, loss, reason, affect & Miscellaneous \\ \hline
        thereto, reaction, chaudhary, manoj, true & Miscellaneous \\ \hline
    \end{tabular}
\end{adjustbox}
\captionsetup{justification=centering}
\caption{
Topics in the entire TCPD-IPD dataset.
\label{Tab:Global_topics}}
\end{table}

\subsection{Topics in subsets of TCPD-IPD}
\label{ap:quality}

 \begin{table}[!htbp]
    \centering
    \begin{adjustbox}{width=1\linewidth}
      \begin{tabular}{| c | c @{\hspace*{1.5mm}} c | c @{\hspace*{1.5mm}} c | c @{\hspace*{1.5mm}} c |} 
        \hline
        %\toprule
        \multirow{2}{*}{\textbf{Dataset}} & \multicolumn{2}{c|}{\textbf{Coherence}} & \multicolumn{2}{c|}{\textbf{Diversity}} & \multicolumn{2}{c|}{\textbf{Topic Quality}}\\
        & \textbf{LDAseq} & \textbf{D-ETM} & \textbf{LDAseq} & \textbf{D-ETM} & \textbf{LDAseq} & \textbf{D-ETM} \\
        \hline
        %\midrule
        Finance & \textcolor{blue}{0.088} & 0.078 & \textcolor{blue}{0.652} & 0.496 & \textcolor{blue}{0.057} & 0.039 \\
        Railways & \textcolor{blue}{0.129} & 0.094 & \textcolor{blue}{0.686} & 0.558 & \textcolor{blue}{0.088} & 0.052 \\
        Health & \textcolor{blue}{0.103} & 0.096 & \textcolor{blue}{0.617} & 0.571 & \textcolor{blue}{0.064} & 0.055 \\
      \hline
      %\bottomrule
     \end{tabular}
    \end{adjustbox}
    \caption{Topic quality analysis. \label{tab:Comaparision}}
\end{table}

 We have carried out a pilot study on dynamic topic modeling with data subsets for the three ministries (Finance, Railways, and Health and Family Welfare). We set the topic count to 20.  
 We have divided each data subset into year-wise slices, partitioned each slice into train:validation:test as 8:1:1, and run LDAseq and D-ETM on them. Then we calculated the coherence (NPMI),  diversity, and topic quality ($= NPMI \times Diversity$) for each model by averaging them over the time slices. Table \ref{tab:Comaparision} shows that LDAseq performs best and hence we choose it to extract topics from the ministry-specific datasets. 
The topics that appear in Figure   \ref{fig:LDAseq_Finance_full_line} in the main text have been  manually labeled by looking at their top five words. Table  \ref{Tab:topics_labels} shows these topic compositions.
 \begin{table}[!htbp]
\centering
\begin{adjustbox}{width=1\linewidth}
  \begin{tabular}{|c|c|l|} \hline 
        \textbf{Year} & \textbf{Top-5 words in topic} & \textbf{Manual label} \\ \hline
        \multicolumn{3}{|c|}{\textbf{Ministry of Finance}} \\ \hline
        1999 & rate, growth, cent, increase, year & \multirow{3}{2cm}{\textit{Economical growth}} \\ \cline{1-2}
        
        2009 & rate, cent, growth, increase, year &\\ \cline{1-2}
        
        2019 & growth, cent, rate, economy, sector & \\ \hline \hline
        
        1999 & bank, rbi, reserve, issue, guideline & \multirow{3}{2cm}{\textit{Banking}} \\ \cline{1-2}
        
        2009 & bank, rbi, issue, guideline, reserve &\\ \cline{1-2}
        
        2019 & bank, rbi, fraud, issue, transaction & \\ \hline \hline
        
        1999 & project, state, world, development, bank & \multirow{3}{2cm}{\textit{Rural development}} \\ \cline{1-2}
        
        2009 & project, state, development, infrastructure, rural & \\ \cline{1-2}
        
        2019 & project, development, state, infrastructure, fund & \\ \hline \hline
        
        1999 & bank, loan, credit, nabard, state & \multirow{3}{2cm}{\textit{Agricultural loan}} \\ \cline{1-2}
        
        2009 & bank, loan, credit, farmer, year & \\ \cline{1-2}
        
        2019 & loan, bank, farmer, credit, scheme & \\ \hline \hline
        
        1999 & scheme, employee, pension, interest, deposit & \multirow{3}{2cm}{\textit{Pension reforms}} \\ \cline{1-2}
        
        2009 & scheme, pension, fund, deposit, interest & \\ \cline{1-2}
        
        2019 & scheme, account, pension, state, pradhan\_mantri & \\ \hline 
        
        %%%%%%%%%%%%%%%%%%%%%%%%%%%%%%%%%%%%%%%%%%%%%%%%%%%%%%%%%%%%%%%%%%%%%%%%%%%%%
        
        \multicolumn{3}{|c|}{\textbf{Ministry of Railways}} \\ \hline
        1999 & work, complete, progress, project, line & \multirow{3}{2cm}{\textit{Railway project}} \\ \cline{1-2}
        
        2009 & work, complete, section, line, gauge\_conversion & \\ \cline{1-2}
        
        2019 & work, section, complete, line, gauge\_conversion & \\ \hline \hline
        
        1999 & freight, traffic, passenger, good, ticket & \multirow{3}{2cm}{\textit{Ticket price}} \\ \cline{1-2}
        
        2009 & ticket, passenger, freight, increase, scheme & \\ \cline{1-2}
        
        2019 & passenger, ticket, fare, freight, train & \\ \hline \hline
        
        1999 & system, safety, track, committee, report & \multirow{3}{2cm}{\textit{Railway safety}} \\ \cline{1-2}
        
        2009 & system, track, safety, committee, signal & \\ \cline{1-2}
        
        2019 & system, track, safety, train, committee & \\ \hline \hline
        
        1999 & project, corporation, rail, development, company & \multirow{3}{2cm}{\textit{Infrastructure development}} \\ \cline{1-2}
        
        2009 & project, development, corridor, rail, identify & \\ \cline{1-2}
        
        2019 & development, project, corridor, rail, high\_speed & \\ \hline \hline
        
        1999 & station, facility, provide, platform, provision & \multirow{3}{2cm}{\textit{Passenger amenity}} \\ \cline{1-2}
        
        2009 & station, facility, provide, platform, work & \\ \cline{1-2}
        
        2019 & station, provide, facility, platform, scheme & \\ \hline

        %%%%%%%%%%%%%%%%%%%%%%%%%%%%%%%%%%%%%%%%%%%%%%%%%%%%%%%%%%%%%%%%%%%%%%%%%%%%%
        \multicolumn{3}{|c|}{\textbf{Ministry of Health and Family Welfare}} \\ \hline
        2000 & research, institute, study, council, develop & \multirow{3}{2cm}{\textit{Medical research}} \\ \cline{1-2}
        2009 & research, study, clinical\_trial, institute, council & \\ \cline{1-2}
        2019 & research, medical, study, clinical\_trial, council & \\ \hline \hline
        
        2000 & child, population, programme, national, reproductive & \multirow{3}{2cm}{\textit{Woman \& child healthcare}} \\ \cline{1-2}

        2009 & child, programme, national, care, woman & \\ \cline{1-2}

        2019 & child, care, woman, national, provide & \\ \hline \hline
        
        2000 & disease, malaria, control, case, death & \multirow{3}{2cm}{\textit{Communicable diseases}} \\ \cline{1-2}

        2009 & disease, control, case, report, malaria & \\ \cline{1-2}

        2019 & patient, treatment, provide, free, scheme & \\ \hline \hline
        
        2000 & centre, care, area, service, rural & \multirow{3}{2cm}{\textit{Rural Medical Infrastructure}} \\ \cline{1-2}

        2009 & patient, treatment, provide, free, hospital & \\ \cline{1-2}

        2019 & national, public, healthcare, provide, include & \\ \hline \hline
        
        2000 & patient, treatment, provide, hospital, free & \multirow{3}{2cm}{\textit{Financial aid}} \\ \cline{1-2}

        2009 & project, crore, fund, expenditure, cost & \\ \cline{1-2}

        2019 & project, fund, crore, cost, completion & \\ \hline %\hline
    \end{tabular}
\end{adjustbox}
\captionsetup{justification=centering}
\caption{
Temporal topics for each ministry in the TCPD-IPD dataset.
\label{Tab:topics_labels}}
\end{table}

\subsection{Keyword extraction for gender and caste-related discussions}
\label{ap:bias}
\begin{table}[!htbp]
\centering
\begin{adjustbox}{width=\linewidth}
  \begin{tabularx}{\linewidth}{|c|L|} \hline 
    \textbf{Word} & \textbf{Top 20 neighbors} \\ \hline
    gender & \textbf{gender}, \textbf{gender\_equality}, \textbf{gender\_disparity}, \textbf{women}, \textbf{woman}, \textbf{gender\_sensitivity}, \textbf{gender\_sensitization}, \textbf{gender\_gap}, \textbf{gender\_parity}, \textbf{girl}, \textbf{gender\_sensitive}, \textbf{female\_literacy}, \textbf{sex}, \textbf{male\_female}, \textbf{disparity}, \textbf{child\_sex}, \textbf{child}, \textbf{girl\_child}, \textbf{literacy\_rate}, \textbf{sex\_selective} \\ \hline
    
    caste & \textbf{caste}, \textbf{tribe}, \textbf{scs}, \textbf{obcs}, \textbf{obc}, \textit{schedule}, \textbf{caste\_tribe}, \textbf{scheduled\_tribe}, \textbf{dalit}, \textbf{social\_justice}, \textbf{scs\_sts}, \textbf{scheduled\_caste}, \textit{vijay\_sampla}, \textit{empowerment\_napoleon}, \textit{belong}, \textit{subbulakshmi\_jagadeesan}, \textbf{atrocity}, \textbf{minority}, \textit{pal\_gurjar}, \textit{empowerment\_smt} \\ \hline
    \end{tabularx}
\end{adjustbox}
\captionsetup{justification=centering}
\caption{Top 20 neighbors (using \texttt{Word2Vec}) for each keyword. 
\label{Tab:keywords}}
\end{table}
To generate keywords for the analysis of gender and caste-related discussions, we have applied SkipGram, which is one of the models used in the neural network-based word2vec algorithm to generate word embeddings \cite{mikolov2013distributed}. We have run SkipGram on the entire TCPD-IPD dataset and taken the top twenty neighbors (based on the cosine similarity of the generated word vectors) of each of the keywords `gender' and `caste'. The words are shown in Table \ref{Tab:keywords}. Then, we counted the documents that contain those  words. We have ignored the words shown in italics in the table as they introduced many irrelevant documents into the count.

\subsection{Explanation of certain terms}
Words in the main text that are difficult to understand outside the Indian context are explained below. 
\begin{enumerate}
    \itemsep0.2em
    \item `npci': National Payments Corporation of India, created by the Reserve Bank of India under the Ministry of Finance, to enable digital payments and settlement systems in India.
    \item `nabard': National Bank for Agriculture and Rural Development, operating under the Ministry of Finance. It regulates the institutions that supply financial help to rural society. 
    \item Kisan Credit Card (KCC): Farmers' Credit Card.
    \item `apy': Atal Pension Yojana. `Yojana' means scheme.
    \item `dbt': Direct Benefit Transfer.
    \item `jan\_dhan': Pradhan Mantri Jan Dhan Yojana. Translates to `Prime Minister's People's Wealth Scheme'.
    \item `rail\_vikas': Railway Vikas Nigam Limited is owned by the Ministry of Railways involved in building rail infrastructure.
    \item `irsdc': Indian Railway Station Development Corporation.
    \item `scs', `obc', `obcs', `scs\_sts', `scheduled\_tribe',  `scheduled\_caste': These words denote historically disadvantaged communities in  India. Scheduled Castes, Scheduled Tribes, and Other Backward Classes are abbreviated as SC, ST, and OBC, respectively.
\end{enumerate}

\section{Bibliographical References}\label{reference}

\section{Language Resource References}
\label{lr:ref}

\end{document}